# A Distance for HMMs based on Aggregated Wasserstein Metric and State Registration

Yukun Chen[†], Jianbo Ye[†], and Jia Li[‡]

†College of Information Sciences and Technology, ‡Department of Statistics
Pennsylvania State University, University Park, USA
{yzc147,jxy198,jol2}@psu.edu

**Abstract.** We propose a framework, named *Aggregated Wasserstein*, for computing a dissimilarity measure or distance between two Hidden Markov Models with state conditional distributions being Gaussian. For such HMMs, the marginal distribution at any time spot follows a Gaussian mixture distribution, a fact exploited to softly match, aka register, the states in two HMMs. We refer to such HMMs as Gaussian mixture model-HMM (GMM-HMM). The registration of states is inspired by the intrinsic relationship of optimal transport and the Wasserstein metric between distributions. Specifically, the components of the marginal GMMs are matched by solving an optimal transport problem where the cost between components is the Wasserstein metric for Gaussian distributions. The solution of the optimization problem is a fast approximation to the Wasserstein metric between two GMMs. The new Aggregated Wasserstein distance is a semi-metric and can be computed without generating Monte Carlo samples. It is invariant to relabeling or permutation of the states. This distance quantifies the dissimilarity of GMM-HMMs by measuring both the difference between the two marginal GMMs and the difference between the two transition matrices. Our new distance is tested on the tasks of retrieval and classification of time series. Experiments on both synthetic data and real data have demonstrated its advantages in terms of accuracy as well as efficiency in comparison with existing distances based on the Kullback-Leibler divergence.

## 1 Introduction

A hidden Markov model (HMM) with Gaussian emission distributions for any given state is a widely used stochastic model for time series of vectors residing in an Euclidean space. It has been massively used in the pattern recognition literature, such as acoustic signal processing (e.g. [1,2,3,4,5,6]) and computer vision (e.g. [7,8,9,10]) for modeling spatial-temporal dependencies in data. We refer to such an HMM as Gaussian mixture model-HMM (GMM-HMM) to stress the fact that the marginal distribution of the vector at any time spot follows a Gaussian mixture distribution. Our new distance for HMMs exploits heavily the GMM marginal distribution, which is the major reason we use the terminology GMM-HMM. We are aware that in some literature, Gaussian mixture HMM is used to mean an HMM with state conditional distributions being Gaussian mixtures rather than a single Gaussian distribution. This more general form of HMM is equivalent to an HMM containing an enlarged set of states with single Gaussian distributions. Hence, it poses no particular difficulty for our proposed framework. More detailed remarks are given in Section 6.

A long-pursued question is how to quantitatively compare two sequences based on the parametric representations of the GMM-HMMs estimated from them respectively. The GMM-HMM parameters lie on a non-linear manifold. Thus a simple Euclidean distance on the parameters is not proper. As argued in the literature (e.g. [11,12]), directly comparing HMM in terms of the parameters is non-trivial, partly due to the *identifiability* issue of parameters in a mixture model. Specifically, a mixture model can only be estimated up to the permutation of states. Different components in a mixture model are actually unordered even though labels are assigned to them, the permutation of labels having no effect on the likelihood of the model. Some earlier solutions do not principally tackle the parameter identifiability issue and simply assume the components are already aligned based on whatever labels given to them [13]. Other more sophisticated solutions sidestep the issue to use model independent statistics including the KL divergence [14,15] and probability product kernels [16,17]. Those statistics usually cannot be computed easily, requiring Monte Carlo samples or the original sequences [12,18], which can be viewed as one source of Monte Carlo samples.

Sometimes approaches that use the original sequence data may give more reliable results than the Monte Carlo approaches. Yet such approaches require that the original sequences are instantly accessible at the phase of data analysis. Imagine a setting where large volumes of data are collected across different sites. Due to the communication constraints or the sheer size of data, it is possible that one cannot transmit all data to a single site. We may have to work on a distributed platform. The models are estimated at multiple sites; and only the models (much compressed information from the original data) are transmitted to a central site. This raises the need of approaches requiring only the model parameters. Existing methods using only the model parameters typically rely on Monte Carlo sampling (e.g. KL-D based methods [14]) to calculate certain log-likelihood statistics. However, the rate of convergence in estimating the log-likelihoods is $O\left(\left(\frac{1}{n}\right)^{2/d}\right)$ [19,20], where $n$ is the data size and $d$ the dimension. This can be slow for GMM-HMMs in high dimensions, not to mention the time to generate those samples.

In this paper, we propose a non-simulation parameter-based framework named *Aggregated Wasserstein* to compute the distance between GMM-HMMs. To address the state identifiability issue, the framework first solves a registration matrix between the states of two GMM-HMMs according to an optimization criterion. The optimization problem is essentially a fast approximation to the Wasserstein metric between two marginal GMMs. Once the registration matrix is obtained, we compute separately the difference between the two marginal

GMMs and the difference between two transition matrices. Finally, we combine the two parts by a weighted sum. The weight can be cast as a trade-off factor balancing the importance between differentiating spatial geometries and stochastic dynamics of two GMM-HMMs.

For an improved estimation of the state registration, we also propose a second approach to calculate the registration matrix based on Monte Carlo sampling. The second approach overcomes certain limitations of the first approach, but at the cost of being more computationally expensive. The second method relies on estimating a mixture weight vector of a special mixture model (explained in our paper), whose rate of convergence is asymptotically $O\left(\sqrt{\frac{\log n}{n}}\right)$ — much faster than the rate of computing log-likelihood based statistics in high dimensions.

We investigate our geometry-driven methods in real world tasks and compare them with the KL divergence-type methods. Practical advantages of our approach have been demonstrated in real applications. By experiments on synthetic data, we also make effort to discover scenarios when our proposed methods outperform the others.

**Our contributions.** We develop a parameter-based framework with the option of not using simulation for computing a distance between GMM-HMMs. Under such framework, a registration matrix is computed for the states in two HMMs. Two methods have been proposed to compute the registration, resulting in two distances, named *Minimized Aggregated Wasserstein* and *Improved Aggregated Wasserstein*. Both distances are experimentally validated to be robust and effective, often outperform KL divergence-based methods in practice.

The rest of the paper is organized as follows. We introduce notations and preliminaries in Section 2. The main framework for defining the distance is proposed in Section 3. The second approach based on Monte Carlo to compute the registration between two sets of HMM states is described in Section 4. Finally, we investigate the new framework empirically in Section 5 based on synthetic and real data.

## 2 Preliminaries

In Section 2.1, we review GMM-HMM and introduce notations. Next, the definition for Wasserstein distance is provided in Section 2.2, and its difference from the KL divergence in the case of Gaussian is discussed.

### 2.1 Notations and Definitions

Consider a sequence $O_T = \{o_1, o_2, ..., o_T\}$ modeled by a GMM-HMM. Suppose there are $M$ states: $S = \{1, ..., M\}$, a GMM-HMM under the stationary condition assumes the following:

1. Each observation $o_i \in O_T$ is associated with a hidden state $s_i \in S$ governed by a Markov chain (MC).

2. $\mathbf{T}$ is the $M \times M$ transition matrix of the MC $\mathbf{T}_{i,j} \overset{\text{def}}{=} P(s_{t+1} = j|s_t = i)$, $1 \le i, j \le M$ for any $t \in \{1, \ldots, T\}$. The stationary (initial) state probability $\pi = [\pi_1, \pi_2, \ldots, \pi_M]$ satisfies $\pi\mathbf{T} = \pi$ and $\pi\mathbf{1} = 1$.

3. The Gaussian probabilistic emission function $\phi_i(o_t) \overset{\text{def}}{=} P(o_t|s_t = i)$, $i = 1, \ldots, M$, for any $t \in \{1, \ldots, T\}$, is the p.d.f. of the normal distribution $\mathcal{N}(\mu_i, \Sigma_i)$, where $\mu_i, \Sigma_i$ are the mean and covariance of the Gaussian distribution conditioned on state $i$.

In particular, we use $\mathcal{M}(\{\mu_i\}_{i=1}^M, \{\Sigma_i\}_{i=1}^M, \pi)$ to denote the corresponding mixture of $M$ Gaussions ( $\{\phi_1, \phi_2, \ldots, \phi_M\}$ ). $\mathcal{M}$'s prior probability of components, aka the mixture weight, coincides with the respective stationary probability $\pi$, which is determined by $\mathbf{T}$. Therefore, one can summarize the parameters for a stationary GMM-HMM model via $\Lambda$ as $\Lambda(\mathbf{T}, \mathcal{M}) = \Lambda(\mathbf{T}, \{\mu_i\}_{i=1}^M, \{\Sigma_i\}_{i=1}^M)$. In addition, the $i$-th row of the transition matrix $\mathbf{T}$ is denoted by $\mathbf{T}(i, :) \in \mathbb{R}^{1 \times M}$. And the next observation's distribution conditioned on current state $i$ is also a GMM: $\mathcal{M}^{(i)}(\{\mu_i\}_{i=1}^M, \{\Sigma_i\}_{i=1}^M, \mathbf{T}(i, :))$, which we abbreviated as $\mathcal{M}^{(i)}|_{\mathbf{T}(i,:)}$.

## 2.2 The Wasserstein Distance and the Gaussian Case

In probability theory, Wasserstein distance is a geometric distance naturally defined for any two probability measures over a metric space.

**Definition 1 (p-Wasserstein distance).** *Given two probability distribution $f, g$ defined on Euclidean space $\mathbb{R}^d$, the $p$-Wasserstein distance $W_p(\cdot, \cdot)$ between them is given by*

$$W_p(f, g) \overset{\text{def}}{=} \left[ \inf_{\gamma \in \Pi(f,g)} \int_{\mathbb{R}^d \times \mathbb{R}^d} \|\mathbf{x} - \mathbf{y}\|^p d\gamma(\mathbf{x}, \mathbf{y}) \right]^{1/p}, \tag{1}$$

*where $\Pi(f, g)$ is the collection of all distributions on $\mathbb{R}^d \times \mathbb{R}^d$ with marginal $f$ and $g$ on the first and second factors respectively. In particular, the $\Pi(\cdot, \cdot)$ is often called as the coupling set. The $\gamma^* \in \Pi(f, g)$ that takes the infimum in Eq. (1) is called the optimal coupling.*

*Remark 1.* By Hölder inequality, one has $W_p \le W_q$ for any $p \le q < \infty$. In this paper, we focus on the practice of $W_p$ with $0 < p \le 2$.

While Wasserstein distance between two multi-dimensional GMMs is unsolved, it has a closed formula for two Gaussian $\phi_1(\mu_1, \boldsymbol{\Sigma}_1)$ and $\phi_2(\mu_2, \boldsymbol{\Sigma}_2)$ [21] when $p = 2$:

$$W_2(\phi_1, \phi_2)^2 = \|\mu_1 - \mu_2\|^2 + tr\left(\boldsymbol{\Sigma}_1 + \boldsymbol{\Sigma}_2 - 2\left(\boldsymbol{\Sigma}_1^{1/2}\boldsymbol{\Sigma}_2\boldsymbol{\Sigma}_1^{1/2}\right)^{1/2}\right). \tag{2}$$

*Remark 2.* The formula of Wasserstein distance between two Gaussians does not involve the inverse-covariance matrix, thus admits the cases of singularity. In comparison, KL divergence between two Gaussian $KL(\phi_1, \phi_2)$ could go to infinity if the covariance of $\phi_2$ becomes singular.

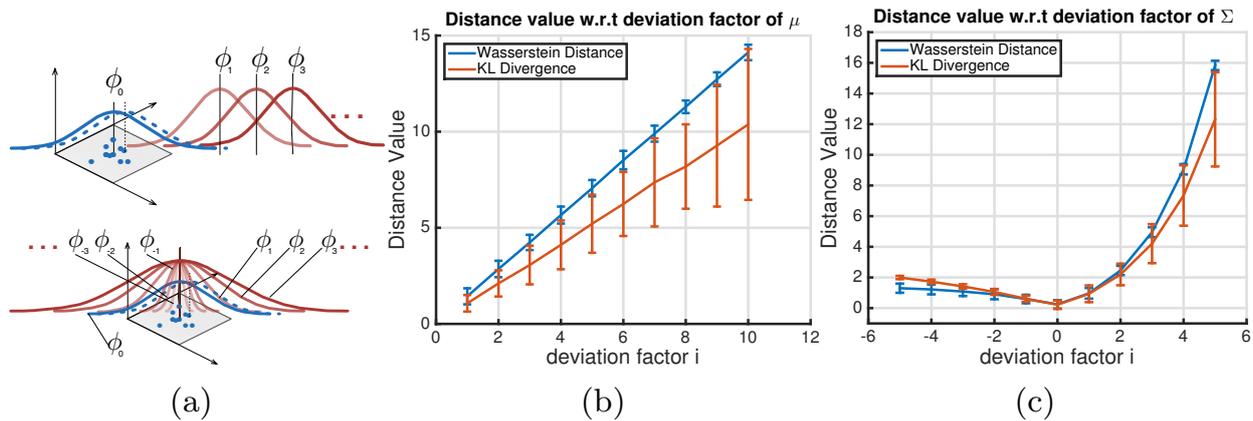

(a)                                          (b)                                          (c)

Fig. 1: (a) Experiment scheme for varying $\mu$ and varying $\Sigma$. A re-estimated $\widehat{\phi}_0$ is denoted as the dashed blue line. (b) (c) Mean estimates of $W_2(\widehat{\phi}_0, \phi_i)$ (blue) and $KL(\widehat{\phi}_0, \phi_i)$ (orange) and their $3\sigma$ confidence intervals w.r.t different Gaussian $\phi_i$. (b) is for varying $\mu$, and (c) is for varying $\Sigma$.

*Remark 3.* The Wasserstein distance could also be more statistically robust than KL divergence by comparing the variance of their estimations. To illustrate this point, we conduct two sets of toy experiments. We sample 100 fixed-size batches of points from pre-selected Gaussian $\phi_0$, re-estimate each batch's Gaussian parameters $\widehat{\phi}_0 = \mathcal{N}(\widehat{\mu}, \widehat{\Sigma}) \approx \phi_0$, and then calculate $W_2(\widehat{\phi}_0, \phi_i)$ and $KL(\widehat{\phi}_0, \phi_i)$, in which $\{\phi_i\}_{i=1}^{10}$ is a sequence of different Gaussians. We construct $\phi_i$ by varying $\mu$ in the first experiment (See Fig. 1(a) upper plot.) and varying $\Sigma$ in the second experiment (See Fig. 1(a) bottom plot.). More detailed experiment setup is explained in Appendix A. Fig. 1 (b) and (c) show the performance of Wasserstein distance and KL divergence on the two toy experiments respectively. Both the estimations and the $3\sigma$ confidence intervals are plotted. It is clear that the Wasserstein distances based on estimated distributions have smaller variance and can overall better differentiate $\{\phi_i\}$.

## 3 The Framework of Aggregated Wasserstein

In this section, we propose a framework to compute the distance between two GMM-HMMs, $\Lambda_1(\mathbf{T}_1, \mathcal{M}_1)$ and $\Lambda_2(\mathbf{T}_2, \mathcal{M}_2)$, where $\mathcal{M}_l$, $l = 1, 2$ are marginal GMMs with pdf $f_l(x) = \sum_{j=1}^{M_l} \pi_{l,j} \phi_{l,j}(x)$ and $\mathbf{T}_1, \mathbf{T}_2$ are the transition matrices of dimension $M_1 \times M_1$ and $M_2 \times M_2$ (recall notations in Section 2). Based on the registration matrix between states in two HMMs, to be described in Section 3.1, the distance between $\Lambda_1$ and $\Lambda_2$ consists of two parts: (1) the difference between $\mathcal{M}_1$ and $\mathcal{M}_2$ (Section 3.2); and (2) the difference between $\mathbf{T}_1$ and $\mathbf{T}_2$ (Section 3.3).

### 3.1   The Registration of States

The registration of states is to build a correspondance between $\Lambda_1$'s states and $\Lambda_2$'s states. In the simplest case (an example is illustrated in Fig 2), if the two marginal GMMs are identical distributions but the states are labeled differently (referred to as permutation of states), the registration should discover the permutation and yield a one-one mapping between the states. We can use a matrix $\mathbf{W} = \{w_{i,j}\} \in \mathbb{R}^{M_1 \times M_2}$ whose elements $w_{i,j} \geq 0$ to encode this registration. In particular, $w_{i,j} = \pi_{1,i}(= \pi_{2,j})$ iff state $i$ in $\Lambda_1$ is registered to state $j$ in $\Lambda_2$. With $\mathbf{W}$ given, through matrix multiplications (details delayed in Section 3.3), the rows and columns of $\mathbf{T}_1$ can be permuted to become identical to $\mathbf{T}_2$.

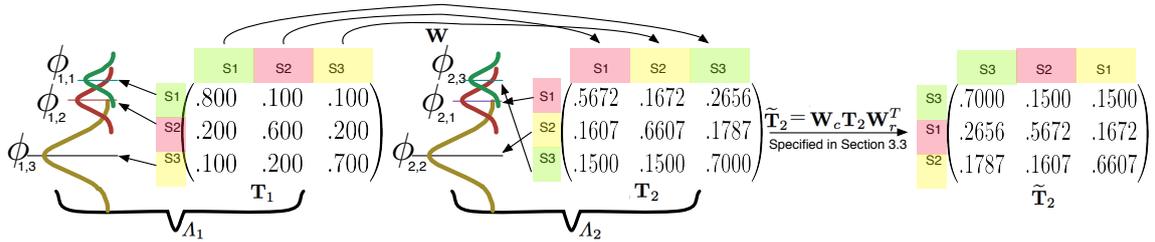

Fig. 2: A simple registration example about how $\mathbf{T}_2$ in $\Lambda_2$ is registered towards $\Lambda_1$ such that it can be compared with $\mathbf{T}_1$ in $\Lambda_1$. For this example, $\mathbf{W}$ encodes a "hard matching" between states in $\Lambda_1$ and $\Lambda_2$

Generally and more commonly, there may exist no state in $\Lambda_2$ having the same emission function as some state in $\Lambda_1$, and the number of states in $\Lambda_1$ may not equal that in $\Lambda_2$. The registration process becomes much more difficult. We resort to the principled optimal transport [22] as a tool to solve this problem and formulate the following optimization problem. Recall Eq. (2) for how to compute $W_2(\phi_{1,i}, \phi_{2,j})$) and let $0 < p \leq 2$, consider

$$\min_{\mathbf{W} \in \Pi(\pi_1, \pi_2)} \sum_{i=1}^{M_1} \sum_{j=1}^{M_2} w_{i,j} W_2(\phi_{1,i}, \phi_{2,j})^p \tag{3}$$

where

$$\Pi(\pi_1, \pi_2) \overset{\text{def}}{=} \Big\{ \mathbf{W} \in \mathbb{R}^{M_1 \times M_2} : \sum_{i=1}^{M_1} w_{i,j} = \pi_{2,j}, j = 1, \dots, M_2;$$
$$\sum_{j=1}^{M_2} w_{i,j} = \pi_{1,i}, i = 1, \dots, M_1; \text{ and } \quad w_{i,j} \geq 0, \forall i, j \Big\} \tag{4}$$

The rationale behind this is that, two states whose emission functions are geometrically close and in similar shape should be more likely to be matched. The solution $\mathbf{W} \in \Pi(\pi_1, \pi_2)$ of the above optimization is called the *registration matrix* between $\Lambda_1$ and $\Lambda_2$. And it will play an important role both in the comparison of marginal GMMs and transition matrices of $\Lambda_1$ and $\Lambda_2$.

The solution of Eq. (3) is an extension of the hard matching between states for the simplest case to the general soft matching when the hard matching is impossible. For the aforementioned simple example (Fig. 2), in which the two Gaussian mixtures are in fact identical thus hard matching is possible, Eq. (3) indeed yields the optimal $\mathbf{W}$ which encodes the correct permutation of states in the two models. In general, there are more than one non-zero elements per row or per column.

### 3.2   The Distance between Two Marginal GMMs

Our aim in this subsection is to quantify the difference between $\Lambda_1$ and $\Lambda_2$'s marginal GMMs $\mathcal{M}_1$ and $\mathcal{M}_2$, whose density functions are $f_1(x) = \sum_{j=1}^{M_1} \pi_{1,j} \phi_{1,j}(x)$ and $f_2(x) = \sum_{j=1}^{M_2} \pi_{2,j} \phi_{2,j}(x)$ respectively.

Given the discussion on the advantages of the Wasserstein metric (especially the Gaussian case) in Section 2, one may ask *why not to use Wasserstein distance $W(\mathcal{M}_1, \mathcal{M}_2)$ directly to measure the dissimilarity between $\mathcal{M}_1, \mathcal{M}_2$?* Unfortunately, there is no closed form formula for GMMs except for the reduced case of single Gaussians. Monte Carlo estimation is usually used. However, similar to the estimation of KL divergence, the Monte Carlo estimation for the Wasserstein distance also suffers from a slow convergence rate. The rate of convergence is as slow as that of KL divergence, i.e., $O\left(\left(\frac{1}{n}\right)^{1/d}\right)$ [23], again posing difficulty in high dimensions. So, instead of estimating the Wasserstein distance itself, we make use of the solved registration matrix $\mathbf{W} \in \Pi(\pi_1, \pi_2)$ and the closed form Wasserstein distance between every pair of Gaussians to quantify the dissimilarity between two marginal GMMs $\mathcal{M}_1$ and $\mathcal{M}_2$:

$$\widetilde{R}_p(\mathcal{M}_1, \mathcal{M}_2; \mathbf{W})^p \overset{\text{def}}{=} \sum_{i=1}^{M_1} \sum_{j=1}^{M_2} w_{i,j} W_2(\phi_{1,i}, \phi_{2,j})^p \tag{5}$$

where $\mathbf{W}$ is the solved registration matrix (from Eq. (3)). Note that registration matrix solved by scheme other than Eq. (3) (e.g. the one we will introduce in Section 4) can also be plugged into this equation. Since we call $\mathbf{W} \in \Pi(\pi_1, \pi_2)$ the registration matrix, we call $\widetilde{R}_p(\mathcal{M}_1, \mathcal{M}_2; \mathbf{W})^p$ the *registered distance* between $\mathcal{M}_1$ and $\mathcal{M}_2$ at $\mathbf{W}$. The motivation for Eq. (5) is that if the matching weights in $\mathbf{W}$ is acceptable, then it seems natural to aggregate the pairwise distances between the Gaussians in the two mixtures through these weights. We will later prove that $\widetilde{R}_p(\mathcal{M}_1, \mathcal{M}_2; \mathbf{W})$ is a semi-metric (Theorem 2). Next, we present Theorem 1 that states this semi-metric as an upper bound on the true Wasserstein metric.

**Theorem 1.** *For any two GMMs $\mathcal{M}_1$ and $\mathcal{M}_2$, let $\widetilde{R}_p(\cdot, \cdot : \mathbf{W})$ be defined as Eq. (5). If $\mathbf{W} \in \Pi(\pi_1, \pi_2)$, we have for $0 < p \leq 2$*

$$\widetilde{R}_p(\mathcal{M}_1, \mathcal{M}_2 : \mathbf{W}) \geq W_p(\mathcal{M}_1, \mathcal{M}_2),$$

*where $W_p(\mathcal{M}_1, \mathcal{M}_2)$ is the true Wasserstein distance between $\mathcal{M}_1$ and $\mathcal{M}_2$ as defined in Eq. (1).*

*Proof.* See appendix B.

For the brevity of notation, if $\mathbf{W}$ is solved from Eq. (3), the resulting distance $\widetilde{R}_p(\mathcal{M}_1, \mathcal{M}_2 : \mathbf{W})$ is denoted by $\widetilde{W}_p(\mathcal{M}_1, \mathcal{M}_2)$.

### 3.3   The Distance between Two Transition Matrices

Given the registration matrix $\mathbf{W}$, our aim in this subsection is to quantify the difference between $\Lambda_1$ and $\Lambda_2$'s transition matrices, $\mathbf{T}_1 \in \mathbb{R}^{M_1 \times M_1}$ and $\mathbf{T}_2 \in \mathbb{R}^{M_2 \times M_2}$. Since the *identifiability* issue is already addressed by the registration matrix $\mathbf{W}$, $\mathbf{T}_2$ can now be registered towards $\mathbf{T}_1$ by the following transform:

$$\widetilde{\mathbf{T}}_2 \stackrel{\text{def}}{=} \mathbf{W}_r \mathbf{T}_2 \mathbf{W}_c^T \in \mathbb{R}^{M_1 \times M_1}, \tag{6}$$

where matrix $\mathbf{W}_r$ and $\mathbf{W}_c$ are row-wise and column-wise normalized $\mathbf{W}$ respectively, a.k.a. $\mathbf{W}_r = \mathrm{diag}^{-1}(\mathbf{W} \cdot \mathbf{1}) \cdot \mathbf{W}$ and $\mathbf{W}_c = \mathbf{W} \cdot \mathrm{diag}^{-1}(\mathbf{1}^T \cdot \mathbf{W})$. A simple example of this process is illustrated in the right part of Fig. 2. Likewise, $\mathbf{T}_1$ can also be registered towards $\mathbf{T}_2$:

$$\widetilde{\mathbf{T}}_1 \stackrel{\text{def}}{=} \mathbf{W}_c^T \mathbf{T}_1 \mathbf{W}_r \in \mathbb{R}^{M_2 \times M_2}. \tag{7}$$

Then, a discrepancy denoted as $D(\mathbf{T}_1, \mathbf{T}_2 : \mathbf{W})$ to measure the dissimilarity of two transition matrices is adopted:

$$D_p(\mathbf{T}_1, \mathbf{T}_2 : \mathbf{W})^p \stackrel{\text{def}}{=} d_T(\mathbf{T}_1, \widetilde{\mathbf{T}}_2)^p + d_T(\mathbf{T}_2, \widetilde{\mathbf{T}}_1)^p \tag{8}$$

where $\widetilde{\mathbf{T}}_1$ and $\widetilde{\mathbf{T}}_2$ are calculated from Eq. (6) and Eq. (7) (with $\mathbf{W}$ given) respectively and

$$d_T(\mathbf{T}_1, \widetilde{\mathbf{T}}_2)^p \stackrel{\text{def}}{=} \sum_{i=1}^{M_1} \pi_{1,i} \widetilde{W}_p \left( \mathcal{M}_1^{(i)}|_{\mathbf{T}_1(i,:)}, \mathcal{M}_1^{(i)}|_{\widetilde{\mathbf{T}}_2(i,:)} \right)^p \tag{9}$$

$$d_T(\mathbf{T}_2, \widetilde{\mathbf{T}}_1)^p \stackrel{\text{def}}{=} \sum_{i=1}^{M_2} \pi_{2,i} \widetilde{W}_p \left( \mathcal{M}_2^{(i)}|_{\mathbf{T}_2(i,:)}, \mathcal{M}_2^{(i)}|_{\widetilde{\mathbf{T}}_1(i,:)} \right)^p \tag{10}$$

We remind that by the notations in Section 2.1, $\mathcal{M}_1^{(i)}|_{\mathbf{T}_1(i,:)}$ is the pdf of the next observation conditioned on the previous state being $i$ (likewise for the other similar terms).

### 3.4   A Semi-metric between GMM-HMMs —- Minimized Aggregated Wasserstein (MAW)

In summary, the dissimilarity between GMM-HMMs $\Lambda_1, \Lambda_2$ comprises two parts: the first is the discrepancy between the marginal GMMs $\mathcal{M}_1, \mathcal{M}_2$, and the second is the discrepancy between two transition matrices after state registration. A weighted sum of these two terms is taken as the final distance. We call this new distance the *Minimized Aggregated Wasserstein (*MAW) between GMM-HMM models. Let $\mathbf{W}$ be solved from Eq. (3).

$$MAW(\Lambda_1, \Lambda_2) \stackrel{\text{def}}{=} (1 - \alpha)\widetilde{R}_p(\mathcal{M}_1, \mathcal{M}_2; \mathbf{W}) + \alpha D_p(\mathbf{T}_1, \mathbf{T}_2 : \mathbf{W}) \tag{11}$$

**Choosing $\alpha$** For the purpose of maximizing the differentiation ability of the distance, $\alpha$ can be determined by maximizing the accuracy obtained by the 1-nearest neighbor classifier on a set of small but representative training GMM-HMMs with ground truth labels.

For clarity, we summarize MAW's computation procedure in Algorithm 1. Theorem 2 states that MAW is a semi-metric. A semi-metric shares all the properties of a true metric (including separation axiom) except for the triangle inequality.

---

**Algorithm 1** Minimized Aggregated Wasserstein (MAW)

---

**Input:** $\Lambda_1 \left( \mathbf{T}_1, \mathcal{M}_1 \left( \{\mu_{1,i}\}_{i=1}^{M_1}, \{\Sigma_{1,i}\}_{i=1}^{M_1} \right) \right),\ \ \Lambda_2 \left( \mathbf{T}_2, \mathcal{M}_2 \left( \{\mu_{2,i}\}_{i=1}^{M_2}, \{\Sigma_{2,i}\}_{i=1}^{M_2} \right) \right)$

**Output:** $MAW(\Lambda_1, \Lambda_2) \in \{0\} \cup \mathbb{R}^+$
 1: Compute registration matrix $\mathbf{W}$ by Eq. (3)
 2: Compute $\widetilde{R}_p(\mathcal{M}_1, \mathcal{M}_2; \mathbf{W})$ by Eq. (5)
 3: Compute $D_p(\mathbf{T}_1, \mathbf{T}_2)$ by Eq. (8), Eq. (9) and Eq. (10)
 4: Compute and return $MAW(\Lambda_1, \Lambda_2)$ defined by Eq. (11).

---

**Theorem 2.** *MAW defined by Eq.* (11) *is a semi-metric for GMM-HMMs if* $0 < \alpha < 1$.

*Proof. See appendix C.*

## 4   Improved State Registration

A clear disadvantage of estimating $\mathbf{W}$ by Eq. (3) and then computing $\widetilde{R}_p(\cdot, \cdot)$ by Eq. (5) is that $\mathbf{W}$ could be sensitive to the parametrization of GMMs. Two GMMs whose distributions are close can be parameterized very differently, especially when the components are not well separated, leading to $\widetilde{W}$ substantially larger than the true Wasserstein metric. In contrast, the real Wasserstein metric $W$ only depends on the underlying distributions, and thus does not suffer from the artifacts caused by the GMM parameterization. In this section, we introduce an improved approach based on Monte Carlo to calculate the registration matrix $\mathbf{W}$ for two GMMs, which can approximate the true Wasserstein metric more accurately than the method specified in Section 3.1.

Suppose the Wasserstein distance between two GMMs $\mathcal{M}_1$ and $\mathcal{M}_2$ are presolved such that the inference for their optimal coupling $\gamma^*$ (referring to Definition 1. ) is at hand. We define a new state registration matrix by

$$\mathbf{W}^* = \int \pi(\mathbf{x}; \mathcal{M}_1)^T \cdot \pi(\mathbf{y}; \mathcal{M}_2) d\gamma^*(\mathbf{x}, \mathbf{y}), \qquad (12)$$

where $\pi(\mathbf{x}; \cdot)$ (a column vector) denotes the posterior mixture component probabilities at point $\mathbf{x}$ inferred from a provided GMM. In Appendix D, we provide mathematical properties of $\mathbf{W}^*$, and show that $\gamma^*$ is also a special mixture model with $M_1 \times M_2$ components whose mixture weights are actually given by vec($\mathbf{W}^*$). Hence a Monte Carlo method to estimate $\mathbf{W}^*$ is hereby given. Two sets ($\{\mathbf{x}_1, \ldots, \mathbf{x}_n\}$ and $\{\mathbf{y}_1, \ldots, \mathbf{y}_n\}$) of equal size i.i.d. samples are generated from $\mathcal{M}_1$ and $\mathcal{M}_2$ respectively. The $\mathbf{W}^*$ is then empirically estimated by

$$\widetilde{\mathbf{W}}_n^* \overset{\text{def}}{=} [\pi(\mathbf{x}_1; \mathcal{M}_1), \ldots, \pi(\mathbf{x}_n; \mathcal{M}_1)] \cdot \Pi_n \cdot [\pi(\mathbf{y}_1; \mathcal{M}_2), \ldots, \pi(\mathbf{y}_n; \mathcal{M}_2)]^T, \quad (13)$$

where $\Pi_n \in \mathbb{R}^{n \times n}$ is the $p$-th optimal coupling solved for the two samples (essentially a permutation matrix). In practice, we use Sinkhorn algorithm to approximately solve for the optimal coupling [24]. $\widetilde{\mathbf{W}}_n^*$ converges to $\mathbf{W}^*$ with probability 1, as $n \to \infty$. Consequently, the **Improved Aggregated Wasserstein** (IAW) is defined similarly as Eq. (11) with a different $\mathbf{W}$ computed from Eq. (13).

*Remark 4 (Convergence Rate).* The estimation of $\mathbf{W}^*$ follows the mixture proportion estimation setting [25,26], whose rate of convergence is $O\left(\sqrt{\dfrac{V_{\widehat{\Pi}} \log n}{n}}\right)$. Here $V_{\widehat{\Pi}} = V_{\widehat{\Pi}}(d, M_1, M_2)$ is the VC dimension of the geometric class induced by the family $\widehat{\Pi}(\mathcal{M}_1, \mathcal{M}_2)$ (See appendix D and [27] for related definitions).

## 5   Experiments

We conduct experiments to quantitatively evaluate the proposed MAW and IAW. In particular, we set $p = 1$. Our comparison baseline is KL based distance [14] since it is the most widely used one (e.g. [12],[11]). In Section 5.1, we use synthetic data to evaluate the sensitivity of MAW and IAW to the perturbation of $\mu$, $\Sigma$, and $\mathbf{T}$. Similar synthetic experiments have been done in related work (e.g. [11]). In Section 5.2, we compare MAW and IAW with KL using the Mocap data under both retrieval and classification settings.

### 5.1   Evaluation of Sensitivity to the Perturbation of Parameters.

Three sets of experiments are conducted to evaluate MAW and IAW's sensitivity to the perturbation of GMM-HMM parameters ($\{\mu_j\}_{j=1}^M$, $\{\Sigma\}_{j=1}^M$, and $\mathbf{T}$) respectively. In each set of experiments, we have five pre-defined 2-state GMM-HMM models $\left\{ \Lambda_j \left( \{\mu_{i,j}\}_{j=1}^2, \{\Sigma_{i,j}\}_{j=1}^2, \mathbf{T}_i \right) \right\}_{i=1}^5$, among which the only difference is GMM means $\{\mu_{i,1}, \mu_{i,2}\}$, GMM covariances $\{\Sigma_{i,1}, \Sigma_{i,2}\}$, or transition matrices $\mathbf{T}_i$. For example, in the 1st experiment, we perturb $\{\mu_{i,1}, \mu_{i,2}\}$ by setting the 5 GMM-HMM's $\{\mu_{i,1}, \mu_{i,2}\}_{i=1}^5$ to be

$$\left\{ \left\{ \begin{pmatrix} 2 + i\Delta\mu \\ 2 + i\Delta\mu \end{pmatrix}, \begin{pmatrix} 5 + i\Delta\mu\cdot \\ 5 + i\Delta\mu\cdot \end{pmatrix} \right\} | i = 1, 2, 3, 4, 5 \right\} \quad (14)$$

respectively. $\{\Sigma_{i,1}, \Sigma_{i,2}\}$ of them are all set to be the same: $\left\{ \begin{pmatrix} 1 & 0 \\ 0 & 1 \end{pmatrix}, \begin{pmatrix} 1 & 0 \\ 0 & 1 \end{pmatrix} \right\}$. And the transition matrix of them are also the same: $\begin{pmatrix} 0.8 & 0.2 \\ 0.2 & 0.8 \end{pmatrix}$. $\Delta\mu$ here is a parameter to control the difference between the 5 models. The smaller the value, the 5 models are more similar to each other and the retrieval will be more challenging. We choose $\Delta\mu$ to be $0.2, 0.4, 0.6$, and compare KL, MAW and IAW under each setting. Please refer to Table 1 for detailed experiment setup for the other two experiments. For each model of the five, 10 sequences of dimension 2 and of length 100 are generated. These ten sequences with their later estimated models form a single class. Therefore in total, we have 50 sequences belonging to 5 classes. During the evaluation, we re-estimate each sequence's GMM-HMM parameters using the well known Baum-Welch Algorithm. Then for each sequence's estimated model, we treat it as a query to retrieve other sequences' models using KL, MAW and IAW respectively. The precision recall plot for the retrieval are shown in Fig. 3.

Table 1: Summary of the parameters setup for parameter perturbation experiments. $rand(2)$ here means random matrix of dimension 2 by 2. $Dirichlet(\boldsymbol{x})$ here means generating samples from Direchlet distribution with parameter $\boldsymbol{x}$.

| Exp. index | deviation step | $\boldsymbol{\mu}$ | $\boldsymbol{\Sigma}$ | $\mathbf{T}$ |
|---|---|---|---|---|
| 1 | $\Delta\mu = 0.2$, $0.4, 0.6$ | $\left\{ \begin{pmatrix} 2+i\Delta\mu \\ 2+i\Delta\mu \end{pmatrix}, \begin{pmatrix} 5+i\Delta\mu \\ 5+i\Delta\mu \end{pmatrix} \right\}$ $\lvert i = 1,2,3,4,5 \rvert$ | $\left\{ \begin{pmatrix} 1 & 0 \\ 0 & 1 \end{pmatrix}, \begin{pmatrix} 1 & 0 \\ 0 & 1 \end{pmatrix} \right\}$ | $\begin{pmatrix} 0.8 & 0.2 \\ 0.2 & 0.8 \end{pmatrix}$ |
| 2 | $\Delta\sigma = 0.2$, $0.4, 0.6$ | $\left\{ \begin{pmatrix} 2 \\ 2 \end{pmatrix}, \begin{pmatrix} 5 \\ 5 \end{pmatrix} \right\}$ | $\{\{0.2 \cdot exp(i\Delta\sigma \cdot \mathbf{S}),$ $0.2 \cdot exp(i\Delta\sigma \cdot \mathbf{S})\} \lvert$ $i = 1,2,3,4,5\},$ $\mathbf{S} = rand(2)$ | $\begin{pmatrix} 0.8 & 0.2 \\ 0.2 & 0.8 \end{pmatrix}$ |
| 3 | $\Delta t = 0.2$, $0.4, 0.6$ | $\left\{ \begin{pmatrix} 2 \\ 2 \end{pmatrix}, \begin{pmatrix} 5 \\ 5 \end{pmatrix} \right\}$ | $\left\{ \begin{pmatrix} 1 & 0 \\ 0 & 1 \end{pmatrix}, \begin{pmatrix} 1 & 0 \\ 0 & 1 \end{pmatrix} \right\}$ | $\{\Delta t \cdot \mathbf{S} + (1-\Delta t) \cdot \mathbf{T}_i \lvert$ $\mathbf{T}_i[j,:] \sim Dirichlet(10 \cdot \mathbf{S}[j,:])$ $i = 1,2,3,4,5\}, \mathbf{S} = \begin{pmatrix} 0.8 & 0.2 \\ 0.2 & 0.8 \end{pmatrix}$ |

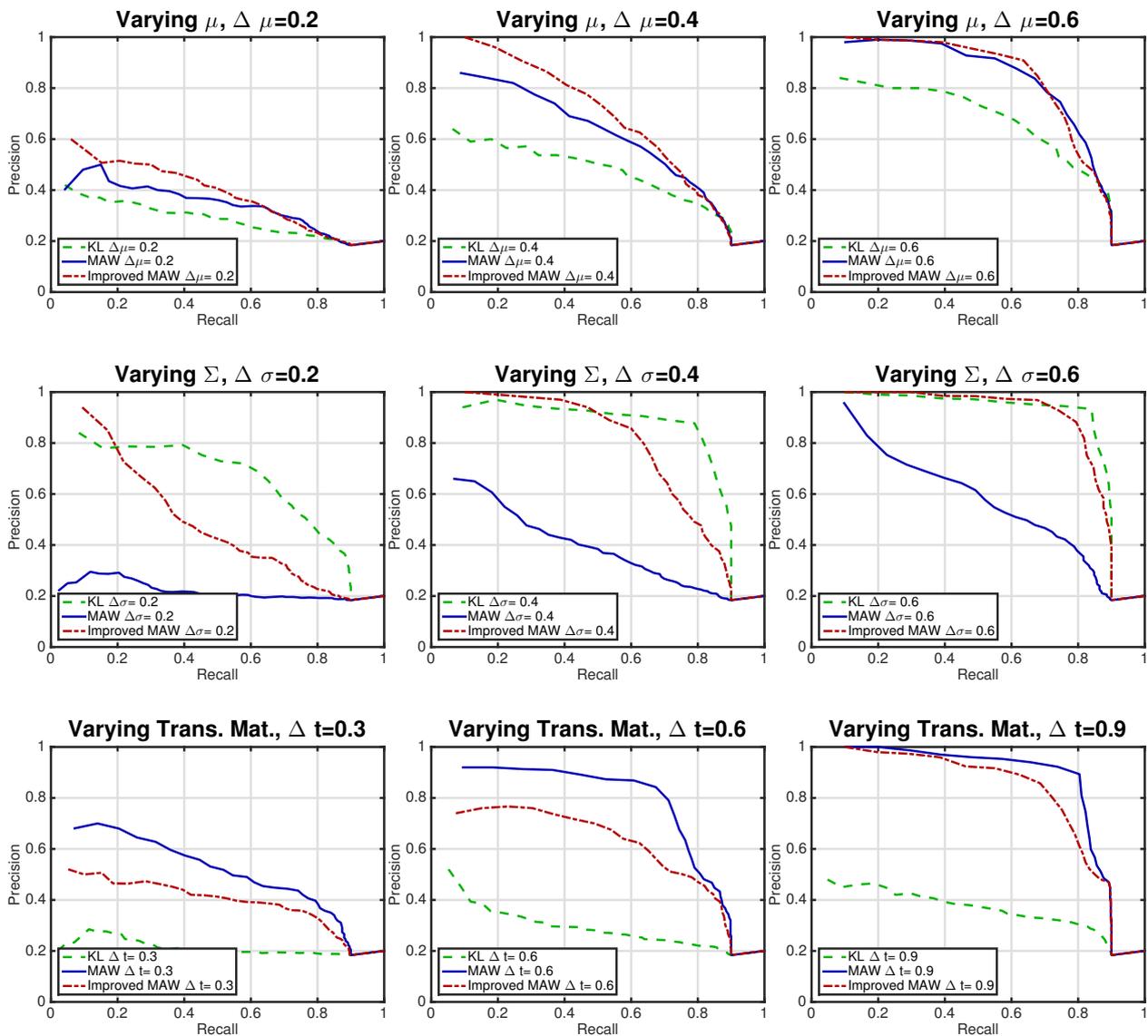

Fig. 3: Precision-recall plot for the study to compare KL, MAW and IAW's sensitivity to the perturbation of GMM-HMM's parameters.

From the first row and third row of Fig. 3, the experiments consistently show that MAW and IAW perform better than KL to differentiate the perturbation of $\{\mu_j\}_{j=1}^{M}$ and $\mathbf{T}$. From the second row, we can see that for the task of differentiating perturbation of $\{\Sigma\}_{j=1}^{M}$, KL performs better than IAW, and IAW performs better than MAW. But for the less challenging case, IAW has comparable performance. Note that if we only care about the nearest neighbor query, IAW actually performs better than KL under the perturbation of $\{\Sigma\}_{j=1}^{M}$. The computation time for MAW and IAW based on our MATLAB implementation are 10ms and 24ms (per distance) respectively and that for KL-D based on our C implementation is 5ms (per distance).[1]

## 5.2   Motion Capture Data

In this section, we use Carnegie Mellon Motion Capture Dataset (Mocap) to evaluate MAW and IAW and make comparison with KL based approach which [18] takes. To improve the stability of evaluation, we only select motion categories 1) whose sequences contain only 1 motion, and 2) which contain more than 20 sequences. In total, there are 7 motion categories, i.e. *Alaskan vacation*, *Jump*, *Story*, *clean*, *salsa dance*, *walk*, and *walk on uneven terrain* that meet this criterion and they contain a total of 337 motion sequences. Since the sequence data is of high dimension (62), following the practice of [18], we split the 62 dimension data to 6 joint-groups[2]. And we conduct both Motion Retrieval based on individual joint-group and Motion Classification using Adaboost on all joint groups. The details of the experiments are specified as follows.

**Motion Retrieval** For each motion time series, we first estimate a 3 state GMM-HMM for each joint-group. Then we use it as a query to retrieve GMM-HMMs estimated from other sequences' on the same joint-group data using KL, MAW and IAW respectively. $\alpha$ for MAW and IAW is chosen such that the 1-nearest neighbor classification accuracy on a small set-aside evaluation set is maximized. The precision-recall plot for the motion retrieval is shown in Fig. 4. The curve is averaged over all motion sequences. We can see that MAW and IAW yield consistently better retrieval results on all joints.

---

[1]  Matlab code: `https://github.com/cykustcc/aggregated_wasserstein_hmm`

[2]  $root_{12}$, $head\_neck\_thorax_{12}$, $rbody_{12}$, $lbody_{12}$, $rleg_6$, $lleg_6$. (The subscript number denotes the dimension of the group)

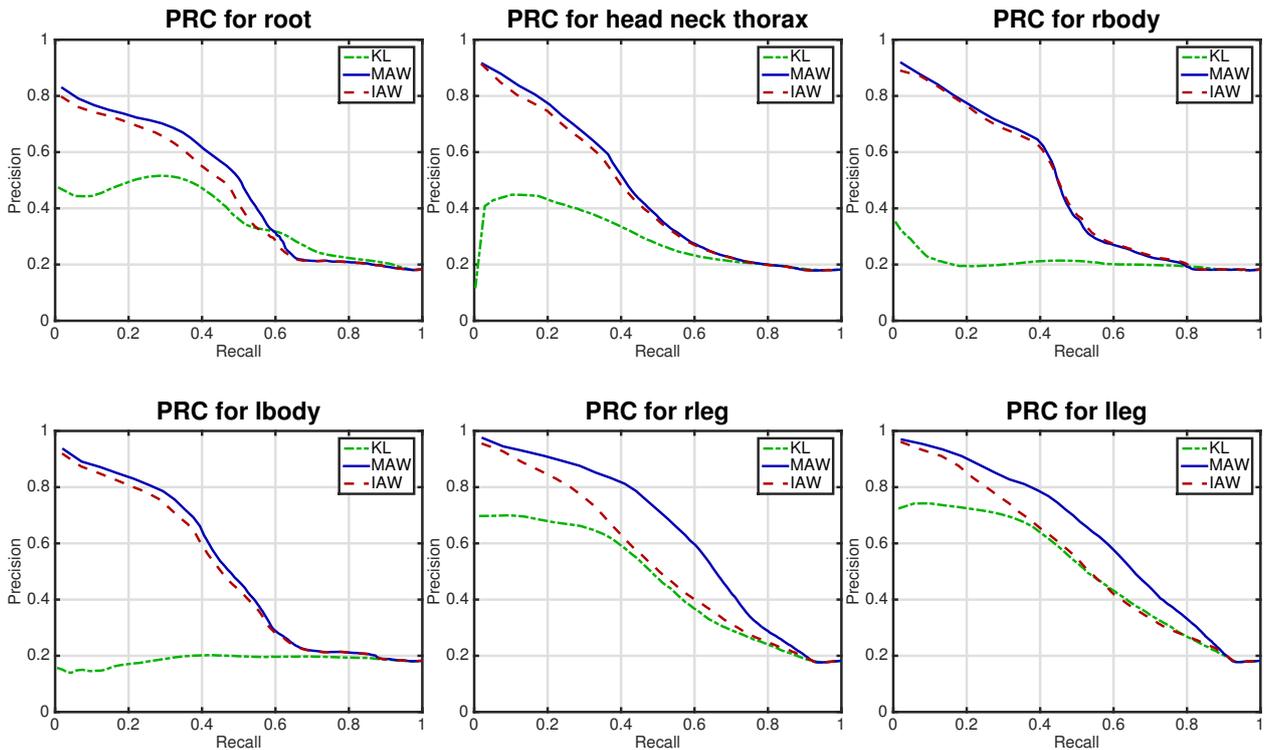

Fig. 4: Precision Recall Plot for Motion Retrieval. The plot for 6 joint-groups, i.e. $root_{12}$, $head\_neck\_thorax_{12}$, $rbody_{12}$, $lbody_{12}$, $rleg_6$, $lleg_6$, are displayed separately.

**Motion Classification** First, we split the 337 motion sequences randomly into two sets, roughly half for training and half for testing. In the training phase, for each of the 7 motion categories, we train one GMM-HMM for each individual joint-group using the training data. For each sequence, we also estimate one GMM-HMM for each individual joint-group. And we compute its distance (either KL, MAW or IAW) to all the GMM-HMMs on the same joint-group data from the 7 motion categories. These distance values are treated as features. The dimension of the feature vector of an individual sequence is thus the number of joint-groups multiplied by 7. Finally, we use Adaboost with depth-one decision trees (essentially, each tree is a one-feature thresholder) to obtain a classification accuracy on the test data. We plot the classification accuracy w.r.t the number of iterations for Adaboost in Fig. 5(a). We also split the original data to 27 different joint groups and run the same experiments again. The results show that under both the 6 joint-group scheme and the 27 joint-group scheme, MAW (85.21% for 6 joint scheme and 95.27% for 27 joint scheme)and IAW (88.17% for 6 joint scheme and 94.08% for 27 joint scheme) achieve considerably better classification accuracy than KL (73.37% for 6 joint scheme and 88.76% for 27 joint scheme ). The confusion matrices are also drawn in Fig. 7 in the supplement.

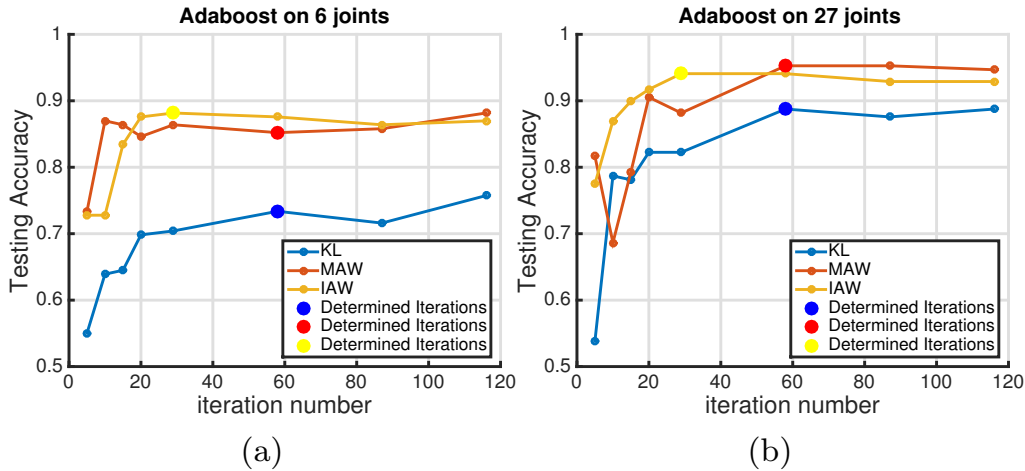

(a)                                    (b)

Fig. 5: Testing accuracies w.r.t iteration number of Adaboost (number of weak classifiers selected). (a) Motion Classification by Adaboost on 6 joints. (b) Motion Classification by Adaboost on 27 joints. The iteration number means the number of features incrementally acquired in Adaboost.

The computation time of Mocap data with 6 joint groups are MAW 21ms, IAW 158ms (1000 samples), and KL-D 8ms (1000 samples). And that of Mocap data with 27 joint groups are MAW 17ms, IAW 160ms (1000 samples), and KL-D 7ms (1000 samples). Again, the MAW and IAW are implemented in MATLAB, and KL-D is implemented in C.

## 6    Discussion and Conclusions

Although we focus on GMM-HMM whose emission function is Gaussian in this paper, the same methodology extends readily to :

1. GMM-HMM whose emission function is GMM but *not* single Gaussian. (Each state with GMM emission function consists of $k$ Gaussians can be split into $k$ states. Our current method can be applied directly then.)
2. Other Hidden Markov Models with non-Gaussian state emission functions, provided that a distance between any two state conditional distributions can be computed. For instance, an HMM with discrete emission distributions can be handled by using the Wasserstein metric between discrete distributions.

In conclusion, we have developed the MAW and IAW distances between GMM-HMMs that are invariant to state permutation. These new distances are computationally efficient, especially MAW. Comparisons with the KL divergence have demonstrated clearly stronger retrieval and classification performance. In future, it is interesting to explore how to reasonably group HMMs into a number of clusters based on our proposed MAW and IAW. The HMM clustering has been studied under the context of KL-D [11], and the clustering under Wasserstein distance has been studied for empirical distributions [28].

**Aknowledgement.** This research is supported by the National Science Foundation under grant number ECCS-1462230.

# A   Toy experiments setup for Fig1 in Section 2.2

To highlight the difference between Wasserstein distance and KL divergence, we conduct two sets of toy experiments to illustrate their statistical natures. We find using Wasserstein distance could be more statistically robust than KL divergence by comparing the variance of their estimations. To illustrate this point, our first toy experiment is shown in Fig. 1 (a) upper figure. First, we sample 100 batches of data, each of size 50, from the pre-selected Gaussian $\phi_0 = \mathcal{N}\left([0,0], \begin{pmatrix} 1,0 \\ 0,1 \end{pmatrix}\right)$. Then, we re-estimate each batch's Gaussian parameters $\widehat{\phi}_0 = \mathcal{N}(\widehat{\mu}, \widehat{\Sigma}) \approx \phi_0$ and calculate $W(\widehat{\phi}_0, \phi_i)$ and $KL(\widehat{\phi}_0, \phi_i)$, in which $\phi_i = \mathcal{N}\left([0.5 \cdot i, 0.5 \cdot i], \begin{pmatrix} 1,0 \\ 0,1 \end{pmatrix}\right)$, $i = 1, ..., 10$ is a sequence of Gaussians, both with closed forms. Ideally, a distance that can consistently differentiate $\phi_i$ by computing its distance to the $\widehat{\phi}$ should have larger value as $i$ grows. Also, its sample deviations of $W_2(\phi_i, \widehat{\phi}_0)$ or $KL(\phi_i, \widehat{\phi}_0)$ should be small enough to not mask out the change from $i$ to $i + 1$. Fig. 1 (b) shows the performance of Wasserstein Distance and KL divergence on this toy experiment. Both the averaged distance to $\phi_i$ and the $3\sigma$ confidence interval are plotted. It's easy to see that Wasserstein Distance is more robust and can differentiate $\phi_i$ better.

Likewise, we also conduct a similar toy experiment by changing $\phi_i$'s variances rather than their means (See Fig. 1 (a) bottom figure). At this time, we set $\phi_i = \mathcal{N}\left([0,0], \exp(0.5 \cdot i) \cdot \begin{pmatrix} 1,0 \\ 0,1 \end{pmatrix}\right)$. The result is plotted in Fig. 1 (c). It shows that KL divergence can be more robust than Wasserstein distance if $\widehat{\phi}_0$ is compared to $\phi_i$ at $i < 0$, but the situation quickly becomes worse at $i \geq 2$. This is due the asymmetric nature of KL divergence. Informally speaking, we conclude from the two toy experiments that estimating $KL(\phi_i, \phi_0)$ can be statistically stable if $\phi_i$ is under the "umbrella" of $\phi_0$, and becomes inaccurate otherwise. On the other hand, Wasserstein distance, as a true metric[22], has consistent accuracy performances across these two settings.

# B   Proof of Theorem 1

*Proof.* We first prove the case for $p = 2$, and then the result for $p \leq 2$ is implied by Hölder inequality. We construct $\gamma \in \Pi(\mathcal{M}_1, \mathcal{M}_2)$ in the following way: Given a $\mathbf{W} \in \Pi(\pi_1, \pi_2)$ and any $\gamma_{z_1, z_2} \in \Pi(\phi_{1, z_1}, \phi_{2, z_2})$ for $z_1 = 1, \ldots, M_1$ and $z_2 = 1, \ldots, M_2$, we let

$$\widetilde{\Pi}(\mathcal{M}_1, \mathcal{M}_2) = \left\{ \gamma \overset{\text{def}}{=} \sum_{z_1=1}^{M_1} \sum_{z_2=1}^{M_2} w_{z_1, z_2} \gamma_{z_1, z_2} \,\middle|\, \mathbf{W} \in \Pi(\pi_1, \pi_2), \right.$$

$$\left. \text{and } \gamma_{i,j} \in \Pi(\phi_{1,i}, \phi_{2,j}), \, i = 1, \ldots, M_1, \, j = 1, \ldots, M_2 \right\} \quad (15)$$

and $\widetilde{R}_2(\cdot, \cdot : \mathbf{W})$ is the exact infimum for all possible $\gamma \in \widetilde{\Pi}(\mathcal{M}_1, \mathcal{M}_2)$, where we see $\widetilde{\Pi}(\mathcal{M}_1, \mathcal{M}_2) \subseteq \Pi(\mathcal{M}_1, \mathcal{M}_2)$. Thus, the inequality is implied.

## C   Proof of Theorem 2

*Proof.* Since Wasserstein Distance is a metric[3],

$$\widetilde{R}_p(\mathcal{M}_1, \mathcal{M}_2; \mathbf{W}) \geq 0 \tag{16a}$$

$$\widetilde{R}_p(\mathcal{M}_1, \mathcal{M}_2; \mathbf{W}) = \widetilde{R}_2(\mathcal{M}_2, \mathcal{M}_1; \mathbf{W}) \tag{16b}$$

$$d_T(\mathbf{T}_1, \widetilde{\mathbf{T}}_2)^p = \sum_{i=1}^{M_1} \pi_{1,i} \widetilde{W}_2 \left( \mathcal{M}_1^{(i)}|_{\mathbf{T}_1(i,:)}, \mathcal{M}_1^{(i)}|_{\widetilde{\mathbf{T}}_2(i,:)} \right)^p \geq 0 \tag{16c}$$

$$d_T(\mathbf{T}_2, \widetilde{\mathbf{T}}_1)^p = \sum_{i=1}^{M_2} \pi_{2,i} \widetilde{W}_2 \left( \mathcal{M}_2^{(i)}|_{\mathbf{T}_2(i,:)}, \mathcal{M}_2^{(i)}|_{\widetilde{\mathbf{T}}_1(i,:)} \right)^p \geq 0 \tag{16d}$$

By Eq. (16a), (16c) and (16d),

$$MAW(\Lambda_1, \Lambda_2) \overset{\text{def}}{=} (1-\alpha)\widetilde{R}_p(\mathcal{M}_1, \mathcal{M}_2; \mathbf{W}) + \alpha D_p(\mathbf{T}_1, \mathbf{T}_2 : \mathbf{W}) \geq 0 \tag{17}$$

And

$$\begin{aligned} D_p(\mathbf{T}_1, \mathbf{T}_2 : \mathbf{W})^p &= d_T(\mathbf{T}_1, \widetilde{\mathbf{T}}_2)^p + d_T(\mathbf{T}_2, \widetilde{\mathbf{T}}_1)^p \\ &= d_T(\mathbf{T}_2, \widetilde{\mathbf{T}}_1)^p + d_T(\mathbf{T}_1, \widetilde{\mathbf{T}}_2)^p = D(\mathbf{T}_2, \mathbf{T}_1 : \mathbf{W})^p \end{aligned} \tag{18}$$

By, Eq. (16b), (18),

$$MAW(\Lambda_1, \Lambda_2) = MAW(\Lambda_2, \Lambda_1) \tag{19}$$

So we have proved MAW is symmetric, greater or equal than zero. And it's obvious that MAW has zero distance between two GMM-HMMs who represent the same distribution. The remaining part is to prove if two GMM-HMMs have zero MAW distance, their distributions are the same.

If $MAW(\Lambda_1, \Lambda_2) = 0$, because $0 < \alpha < 1$ and by Eq. (16a),(16c) and (16d),

$$\widetilde{R}_p(\mathcal{M}_1, \mathcal{M}_2; \mathbf{W}) = 0 \tag{20a}$$

$$D_p(\mathbf{T}_1, \mathbf{T}_2 : \mathbf{W}) = 0 \tag{20b}$$

By Eq. (20a) and the fact that Wasserstein distance for Gaussian is a true metric, $\mathcal{M}_1$ and $\mathcal{M}_2$ should be identical.

By Eq. (16c), (16d), and the fact that Wasserstein distance for Gaussian is a true metric

$$\widetilde{W}_p \left( \mathcal{M}_1^{(i)}|_{\mathbf{T}_1(i,:)}, \mathcal{M}_1^{(i)}|_{\widetilde{\mathbf{T}}_2(i,:)} \right) = 0 \quad \widetilde{W}_p \left( \mathcal{M}_2^{(i)}|_{\mathbf{T}_2(i,:)}, \mathcal{M}_2^{(i)}|_{\widetilde{\mathbf{T}}_1(i,:)} \right) = 0 \tag{21}$$

---

[3] Rachev, Svetlozar T. "The Monge-Kantorovich mass transference problem and its stochastic applications." Theory of Probability & Its Applications 29.4 (1985): 647-676.

That is $\mathcal{M}_1^{(i)}|_{\mathbf{T}_1(i,:)}$, $\mathcal{M}_1^{(i)}|_{\widetilde{\mathbf{T}}_2(i,:)}$ are identical and $\mathcal{M}_2^{(i)}|_{\mathbf{T}_2(i,:)}$, $\mathcal{M}_2^{(i)}|_{\widetilde{\mathbf{T}}_1(i,:)}$ are identical. So $\mathbf{T}_1 = \mathbf{T}_2$. Then $\Lambda_1(\mathcal{M}_1, \mathbf{T}_1)$ and $\Lambda_2(\mathcal{M}_2, \mathbf{T}_2)$ should be identical. So, we have proved $MAW$ is a semi-metric.

Note that by Eq. (16a), (16b) and the fact that $\widetilde{R}_2(\mathcal{M}_1, \mathcal{M}_2; \mathbf{W}) = 0$ iff $\mathcal{M}_1$ and $\mathcal{M}_2$ are identical, we also proved that $\widetilde{R}_2(\mathcal{M}_1, \mathcal{M}_2; \mathbf{W})$ is a metric for GMM. (We mentioned this at Section 3.2)

## D  Property of W*

For the ease of notation, we assume $p = 2$. The proof also applies to any $0 < p \leq 2$ under trivial modification. In Appendix B, we see constructing $\widetilde{\Pi}$ (defined by Eq. (15)) involves a set of strong constraints that $\gamma_{i,j} \in \Pi(\phi_{1,i}, \phi_{2,j})$ for all $i, j$. We try to relax these constraints in this section to derive a different method. Consider measure $\gamma_{i,j} \in \Pi(\tilde{\phi}_{1,i,j}, \tilde{\phi}_{2,i,j})$ where $\tilde{\phi}_{1,i,j}$ and $\tilde{\phi}_{2,i,j}$ are marginals of $\gamma_{i,j}$, treated as parameters of set $\Pi(\tilde{\phi}_{1,i,j}, \tilde{\phi}_{2,i,j})$. In order to ensure $\sum_{z_1=1}^{M_1} \sum_{z_2=2}^{M_2} w_{z_1,z_2} \gamma_{z_1,z_2} \in \Pi(\mathcal{M}_1, \mathcal{M}_2)$, we expose a sufficient condition for parameters $\{\tilde{\phi}_{1,i,j}, \tilde{\phi}_{2,i,j}\}$ as follows:

**Definition 2.** *Given* $\mathbf{W} \in \Pi(\pi_i, \pi_j)$, *we say* $\{\tilde{\phi}_{1,i,j}, \tilde{\phi}_{2,i,j}\}$ *couple with* $\mathbf{W}$ *subject to* $(\mathcal{M}_1, \mathcal{M}_2)$ *if for all* $i = 1, \ldots, M_1$ *and* $j = 1, \ldots, M_2$,

$$\sum_{j=1}^{M_2} \frac{w_{i,j}}{\pi_{1,i}} \tilde{\phi}_{1,i,j} = \phi_{1,i}, \qquad \sum_{i=1}^{M_1} \frac{w_{i,j}}{\pi_{2,j}} \tilde{\phi}_{2,i,j} = \phi_{2,j}. \qquad (22)$$

*We denote these conditions collectively by*

$$\{\tilde{\phi}_{1,i,j}, \tilde{\phi}_{2,i,j}\} \in \Gamma(\mathbf{W} \in \Pi(\pi_i, \pi_j) | \mathcal{M}_1, \mathcal{M}_2). \qquad (23)$$

Alternatively speaking, if $\{\tilde{\phi}_{1,i,j}, \tilde{\phi}_{2,i,j}\}$ couple with $\{w_{i,j}\}$ subject to $(\mathcal{M}_1, \mathcal{M}_2)$, we immediately have

$$\sum_{z_1=1}^{M_1} \sum_{z_2=2}^{M_2} w_{z_1,z_2} \gamma_{z_1,z_2} \in \Pi(\mathcal{M}_1, \mathcal{M}_2),$$

for any $\gamma_{i,j} \in \Pi(\tilde{\phi}_{1,i,j}, \tilde{\phi}_{2,i,j})$, $i = 1, \ldots, M_1$ and $j = 1, \ldots, M_2$. We let

$$\widehat{\Pi}(\mathcal{M}_1, \mathcal{M}_2) \overset{\text{def}}{=} \Bigg\{ \gamma \overset{\text{def}}{=} \sum_{z_1=1}^{M_1} \sum_{z_2=1}^{M_2} w_{z_1,z_2} \gamma_{z_1,z_2} \Bigg| \gamma_{i,j} \in \Pi(\tilde{\phi}_{1,i,j}, \tilde{\phi}_{2,i,j}),$$

$$\{\tilde{\phi}_{1,i,j}, \tilde{\phi}_{2,i,j}\} \in \Gamma(\mathbf{W} | \mathcal{M}_1, \mathcal{M}_2), \mathbf{W} \in \Pi(\pi_1, \pi_2) \Bigg\}. \qquad (24)$$

From the definitions, one can verify that $\widetilde{\Pi}(\mathcal{M}_1, \mathcal{M}_2) \subseteq \widehat{\Pi}(\mathcal{M}_1, \mathcal{M}_2) \subseteq \Pi(\mathcal{M}_1, \mathcal{M}_2)$. Therefore, optimize transportation cost over $\gamma \in \widehat{\Pi}(\mathcal{M}_1, \mathcal{M}_2)$ gives a tighter upper bound of $W(\mathcal{M}_1, \mathcal{M}_2)$ than $\widetilde{W}(\mathcal{M}_1, \mathcal{M}_2)$. Moreover we have

**Theorem 3.** *Following the aforementioned definitions, we have*

$$\widehat{\Pi}(\mathcal{M}_1, \mathcal{M}_2) = \Pi(\mathcal{M}_1, \mathcal{M}_2).$$

*Hence, the optimal coupling in Eq. (1) can be factored as a finite mixture model with $M_1 \cdot M_2$ components, whose proportion vector is $\mathbf{W}^*$: $\mathbf{W}^*$ is taken from the minimizer $\gamma^* = \sum_{i,j} w_{i,j}^* \gamma_{i,j}^* \in \widehat{\Pi}(\mathcal{M}_1, \mathcal{M}_2)$ of the following problem*

$$\inf_{\gamma \in \widehat{\Pi}(\mathcal{M}_1, \mathcal{M}_2)} \int_{\mathbb{R}^d \times \mathbb{R}^d} \|\mathbf{x} - \mathbf{y}\|^2 d\gamma(\mathbf{x}, \mathbf{y}). \tag{25}$$

*Proof.* See appendix E.

*Remark 5.* Theorem 3 illustrates an alternative perspective to interpret any coupling between two Gaussian mixtures (and this perspective can also be generalized to any mixture models): The coupling $\gamma$ of two mixture models can be recast procedurally as firstly dividing each component $\phi_{1,i}, \phi_{2,j}$ into a set of smaller ones $\{\tilde{\phi}_{1,i,j}\}_j, \{\tilde{\phi}_{2,i,j}\}_i$ respectively, whose proportions of mass are $\{w_{i,j}\}$, and then find coupling $\gamma_{i,j}$ for each pair in $\Pi(\tilde{\phi}_{1,i,j}, \tilde{\phi}_{2,i,j})$.

# E   Proof of Theorem 3

To prove this, we only need to show that for any $\gamma \in \Pi(\mathcal{M}_1, \mathcal{M}_2)$, there exist $w_{i,j} \in \Pi(\pi_1, \pi_2)$, $\{\tilde{\phi}_{1,i,j}, \tilde{\phi}_{2,i,j}\} \in \Gamma(\{w_{i,j}\} | \mathcal{M}_1, \mathcal{M}_2)$ and $\gamma_{i,j} \in \Pi(\tilde{\phi}_{1,i,j}, \tilde{\phi}_{2,i,j})$ with $i = 1, \dots, M_1$ and $j = 1, \dots, M_2$ such that

$$\gamma = \sum_{z_1=1}^{M_1} \sum_{z_2=1}^{M_2} w_{z_1,z_2} \gamma_{z_1,z_2}.$$

The constructive proof goes in two steps:

First, given any random variables $(x_1, x_2) \sim \gamma \in \Pi(\mathcal{M}_1, \mathcal{M}_2)$, we can induce component membership random variables $(z_1, z_2)$ by

$$p(z_1, z_2) = \int_{\mathbb{R}^d \times \mathbb{R}^d} p(z_1, z_2 | x_1, x_2) d\gamma(x_1, x_2) \stackrel{\text{def}}{=} w_{z_1, z_2}, \tag{26}$$

where the condition probability is defined multiplicatively by

$$p(z_1, z_2 | x_1, x_2) \stackrel{\text{def}}{=} \frac{\pi_{1,z_1} \phi_{1,z_1}(x_1)}{f_1(x_1)} \cdot \frac{\pi_{2,z_2} \phi_{2,z_2}(x_2)}{f_2(x_2)}. \tag{27}$$

One can verify that $\{w_{i,j}\} \in \Pi(\pi_1, \pi_2)$ by the definition of Eq. (26): for any $i = 1, \ldots, M_1$

$$\sum_{j=1}^{M_2} w_{i,j} = \int_{\mathbb{R}^d \times \mathbb{R}^d} \sum_{j=1}^{M_2} p(z_1 = i, z_2 = j | x_1, x_2) d\gamma(x_1, x_2)$$

$$= \int_{\mathbb{R}^d \times \mathbb{R}^d} \frac{\pi_{1,i} \phi_{1,i}(x_1)}{f_1(x_1)} d\gamma(x_1, x_2)$$

$$= \int_{\mathbb{R}^d} \pi_{1,i} \phi_{1,i}(x_1) dx_1 \text{ (integral out } x_2, \text{ since } \gamma \in \Pi(\mathcal{M}_1, \mathcal{M}_2))$$

$$= \pi_{1,i}.$$

Likewise, $\sum_{i=1}^{M_1} w_{i,j} = \pi_{2,j}$ for any $j = 1, \ldots, M_2$.

Second, consider the conditional measure

$$\gamma(x_1, x_2 | z_1, z_2) = p(z_1, z_2 | x_1, x_2) \gamma(x_1, x_2) / w_{z_1, z_2}$$

(by the Bayes rule), its marginals are

$$d\gamma(x_1 | z_1, z_2) = \frac{1}{w_{z_1, z_2}} \int_{x_2 \in \mathbb{R}^d} p(z_1, z_2 | x_1, x_2) d\gamma(x_1, x_2),$$

$$d\gamma(x_2 | z_1, z_2) = \frac{1}{w_{z_1, z_2}} \int_{x_1 \in \mathbb{R}^d} p(z_1, z_2 | x_1, x_2) d\gamma(x_1, x_2).$$

By definition, we know $\gamma(x_1, x_2 | z_1, z_2) \in \Pi(\gamma(x_1 | z_1, z_2), \gamma(x_2 | z_1, z_2))$. One can validate that $\{\gamma(x_1 | z_1, z_2), \gamma(x_2 | z_1, z_2)\} \in \Gamma(\{w_{i,j}\} | \mathcal{M}_1, \mathcal{M}_2)$: for $z_1 = 1, \ldots, M_1$ and $z_2 = 1, \ldots, M_2$,

$$\sum_{z_2=1}^{M_2} \frac{w_{z_1, z_2}}{\pi_{1, z_1}} d\gamma(x_1 | z_1, z_2) = \int_{x_2 \in \mathbb{R}^d} \sum_{z_2=1}^{M_2} \frac{\phi_{1, z_1}(x_1)}{f_1(x_1)} \cdot \frac{\pi_{2, z_2} \phi_{2, z_2}(x_2)}{f_2(x_2)} d\gamma(x_1, x_2)$$

$$= \int_{x_2 \in \mathbb{R}^d} \frac{\phi_{1, z_1}(x_1)}{f_1(x_1)} d\gamma(x_1, x_2)$$

$$= \phi_{1, z_1}(x_1) dx_1.$$

Likewise, we can show $\sum_{z_1=1}^{M_1} \frac{w_{z_1, z_2}}{\pi_{2, z_2}} d\gamma(x_2 | z_1, z_2) = \phi_{2, z_2}(x_2) dx_2$. Let $\tilde{\phi}_{l,i,j}$ be the p.d.f. of $\gamma(x_l | z_1 = i, z_2 = j)$ and $\gamma_{i,j} \overset{\text{def}}{=} \gamma(x_1, x_2 | z_1 = i, z_2 = j)$, we see that $\gamma \in \widehat{\Pi}(\mathcal{M}_1, \mathcal{M}_2)$. Therefore, $\Pi(\mathcal{M}_1, \mathcal{M}_2) \subseteq \widehat{\Pi}(\mathcal{M}_1, \mathcal{M}_2)$. Combined with the fact that $\widehat{\Pi}(\mathcal{M}_1, \mathcal{M}_2) \subseteq \Pi(\mathcal{M}_1, \mathcal{M}_2)$, the proof is complete.

# F    Additional Results

For experiments in Section 5.1, we also draw graphs similar to Fig. 1 as follows. As we have explained in 5.1, we have 5 HMMs, each of which generates 10 sequences. Specifically, we visualize the mean and 1-$\sigma$ deviation of the distance (MAW, IAW or KL) of the sequences from the first HMM to the sequences from all the five HMMs. Totally we have 9 figures and They can be used to compare the differentiation ability of MAW, IAW and KL from a perspective similar to Fig. 1's. The results again corroborate the conclusion we made after Fig. 3 in Section 5.1.

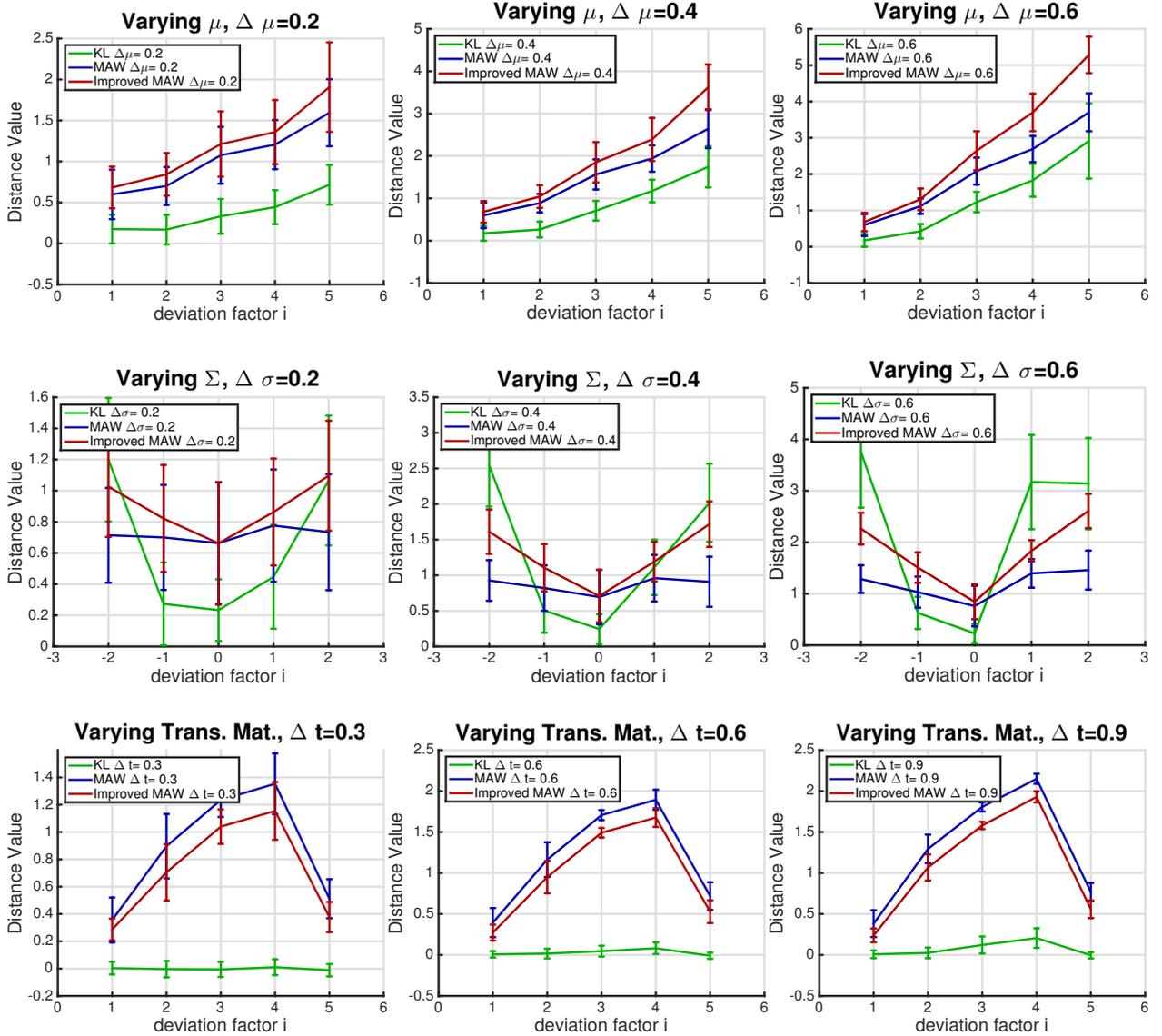

Fig. 6: Comparison of Differentiation Ability. Here only the intervals subject to sample variance are compared w.r.t. the change of their distance values.

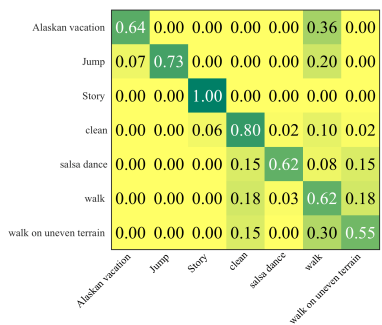
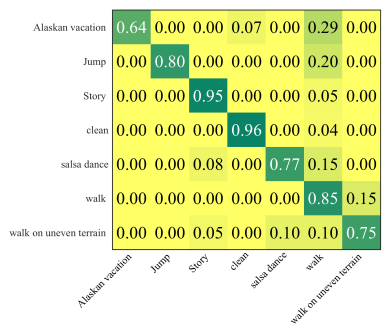
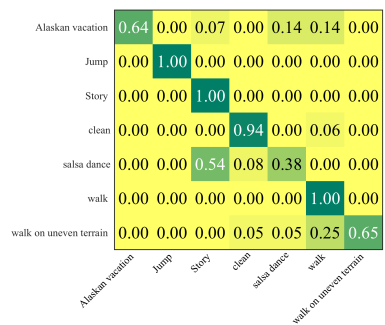

(a)                                              (b)                                              (c)

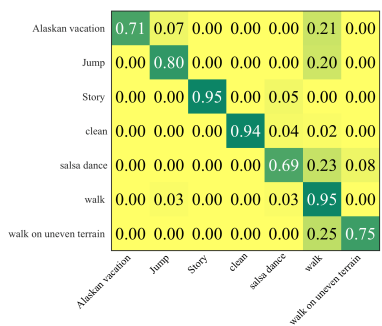
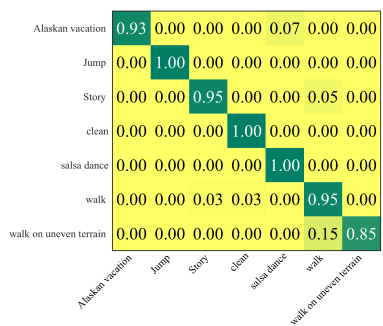
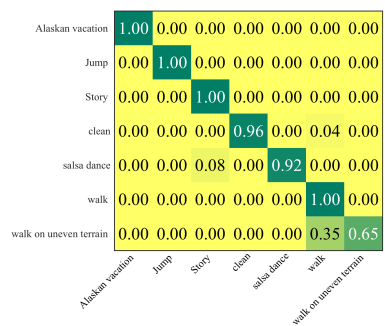

(d)                                              (e)                                              (f)

Fig. 7: (a) 6 joints, KL, corresponds to the blue dot in Fig. 5 (a), (b) 6 joints, MAW, corresponds to the red dot in Fig. 5 (a), (c) 6 joints, IAW, corresponds to the yellow dot in Fig. 5 (a), (d) 27 joints, KL, corresponds to the blue dot in Fig. 5 (b), (e) 27 joints, MAW, corresponds to the red dot in Fig. 5 (b), (f) 27 joints, IAW, corresponds to the yellow dot in Fig. 5 (b),